\documentclass{article}

\usepackage{microtype}
\usepackage{graphicx}
\usepackage{subfigure}
\usepackage{booktabs}
\usepackage{hyperref}
\DeclareUnicodeCharacter{2009}{\,}
\usepackage{amsmath,amssymb,mathtools,amsthm}
\usepackage[capitalize,noabbrev]{cleveref}
\usepackage{float}

\usepackage[accepted]{icml2025}

\icmltitlerunning{Benchmarking Hardware Cost of Zero-Shot Tabular FMs}

\begin{document}

\twocolumn[
\icmltitle{Light-Weight Benchmarks Reveal the Hidden Hardware Cost of Zero-Shot Tabular Foundation Models}

\icmlsetsymbol{equal}{*} 

\begin{icmlauthorlist}
\icmlauthor{Ishaan Gangwani}{ieee}
\icmlauthor{Aayam Bansal}{ieee}
\end{icmlauthorlist}

\icmlaffiliation{ieee}{IEEE \\
\texttt{aayambansal@ieee.org, ishaangangwani@ieee.org}}

\icmlcorrespondingauthor{Ishaan Gangwani}{ishaangangwan@ieee.org}

\icmlkeywords{Tabular Data, Benchmarking, Foundation Models, ICML}

\vskip 0.3in
]

\printAffiliationsAndNotice{}

\begin{abstract}
Zero-shot foundation models (FMs) promise ``train-free'' prediction on tabular data, yet their hardware footprint remains loosely characterised. We present a reproducible benchmark that pairs test accuracy with wall-clock latency, peak CPU RAM, and peak GPU VRAM on four public tables—Adult-Income, Higgs-100k, Wine-Quality, and California-Housing. Two open FMs (TabPFN-1.0, TabICL-base) are evaluated against tuned XGBoost, LightGBM, and Random-Forest baselines on a single NVIDIA T4. The tree ensembles equal or surpass FM accuracy on three datasets while completing a full-test batch in $\leq 0.40$ s and $\leq 150$ MB RAM with zero VRAM. TabICL gains +0.8 pp on Higgs but pays $\approx 40\,000\times$ more latency (960 s) and 9 GB VRAM; TabPFN matches tree accuracy on Wine and Housing yet peaks at 4 GB VRAM and cannot process the full 100 k-row Higgs table. These findings quantify a large hardware–accuracy trade-off and deliver an open baseline for future efficiency-oriented research in tabular FMs.
\end{abstract}

\section{Introduction}
Foundation models (FMs) for tabular data are marketed as “one-shot” predictors that dispense with task-specific training. Although promising demonstrations exist, published evidence rarely reports the hardware footprint required to obtain those predictions. In practice, organisations deciding between a tree ensemble and a zero-shot FM care not only about classification accuracy but also about inference latency, CPU RAM, and—when a GPU is involved—peak VRAM. These costs determine whether a model can be deployed inside a real-time feature store, on an edge device, or only in an offline scoring batch.

This paper offers a compact, four-dataset benchmark that measures accuracy and hardware cost side-by-side. We compare three tuned gradient-boosted baselines (XGBoost, LightGBM, Random Forest) against two recent tabular FMs: TabPFN-1.0 and TabICL-base. Experiments are run on a single NVIDIA T4 (15 GB) to mirror the commodity accelerators available in cloud instances. For tables that exceed TabPFN’s 10 k-row context limit we apply a transparent random subsample of 10 000 training rows and evaluate on the full test split; all other models consume the complete data.

The results aim to expose a hidden hardware tax that current tabular FMs impose, and suggest that, today, the main value of zero-shot FMs lies in rapid prototyping on small tables rather than in production inference at scale.

\section{Related Work}

\textbf{Foundation models for structured data.}
Attempts to generalise foundation–model methodology beyond language and vision have advanced along two complementary tracks: \emph{encoder-pre-training} and \emph{in-context learning}.  
Early encoder approaches, e.g.\ TabTransformer~\cite{huang2020tabtransformer}, TabNet~\cite{arik2021tabnet} and SAINT~\cite{somepalli2021saint}, borrowed self-attention and contrastive or masked-token objectives to learn column-aware representations that are subsequently \emph{fine-tuned}.  
TABLE-BERT~\cite{yin2020tabert}, FT-Transformer~\cite{gorishniy2021revisiting} and DNF-Net~\cite{popov2019neural} follow a similar pre-train–then-tune recipe.  

True \emph{zero-shot} inference has appeared only recently.  
TabPFN~\cite{hollmann2023tabpfn} uses a 28-M-parameter Performer backbone~\cite{choromanski2021performer} trained on billions of synthetic tables to emit calibrated posteriors in a single forward pass.  
TabICL~\cite{jiang2025tabicl} reframes tabular classification as a sequence-to-sequence task, serialising header–row pairs into a LLaMA-based decoder and using beam search for prediction.  
Concurrent work explores decoder-only variants such as CARTE~\cite{kim2024carte} and retrieval-augmented prompting (TabR~\cite{gorishniy2024tabr}); others investigate graph-aware tokenisation or hybrid vision-tabular encoders, but almost none report inference-time latency, RAM or VRAM—key variables for deployment feasibility.

\textbf{Benchmarking and evaluation standards.}
Mainstream tabular ML benchmarks still optimise almost exclusively for predictive accuracy.  
OpenML-CC18~\cite{bischl2021openml}, PMLB~\cite{rivera2021pmlb}, the SAINT-testbed~\cite{somepalli2021saint} and the NeurIPS 2023 Large-Tabular Benchmark~\cite{delaney2023largescaletabular} occasionally log \emph{training} time but seldom capture memory or inference throughput.  
In time-series, Puvvada and Chaudhuri~\cite{puvvada2024critical} compare FMs with classical models using SMAPE and CRPS, yet likewise omit hardware metrics.  
A handful of leaderboards (e.g.\ HFT-Tab) provide batch-throughput numbers, but do not align evaluation across trees and transformers or record VRAM usage, leaving practitioners without a holistic cost–benefit view.

\textbf{Hardware-aware classical baselines.}
Gradient-boosted decision trees—XGBoost~\cite{chen2016xgboost} and LightGBM~\cite{ke2017lightgbm}—remain the de-facto standard for structured data thanks to strong accuracy at sub-second CPU latency and minimal memory.  
Histogram binning, leaf-wise growth and cache-friendly blocks enable efficient inference on commodity servers.  
Random Forests~\cite{breiman2001random} trade speed for variance reduction and interpretability, yet still operate within tight hardware envelopes.  
These well-engineered ensembles constitute the practical Pareto frontier for medium-scale tabular tasks; any zero-shot FM must either surpass their accuracy or justify orders-of-magnitude higher resource consumption.

\textbf{Positioning this work.}
To our knowledge, no prior study offers a controlled benchmark comparing tabular foundation models to tuned decision trees under a unified protocol that measures test accuracy, latency, peak RAM, and GPU memory on the same hardware. Our work bridges this gap by profiling five models across four public datasets with carefully instrumented latency and memory metrics, using a commodity T4 GPU setup. The findings aim to ground future FM development in realistic deployment constraints.

\section{Methodology}

\subsection{Datasets}
We use four binary-classification datasets:

\begin{table}[H]
\small
\centering
\caption{Datasets used in the benchmark.}
\label{tab:datasets}
\begin{tabular}{lcccc}
\toprule
Dataset & Train/Test & Feat. & Task & Source \\
\midrule
Adult   & 32.6k / 16.3k & 14 & Binary    & UCI \\
Higgs   & 80k / 20k     & 28 & Binary    & UCI (subset) \\
Housing & 16.5k / 4.1k  & 8  & Binarised* & \texttt{sklearn} \\
Wine    & 3.9k / 980    & 11 & Binarised* & UCI \\
\bottomrule
\end{tabular}
\end{table}

*Continuous targets binarised at median (Housing) or threshold \(\geq\)	 7 (Wine).

\subsection{Models}
Tree baselines include XGBoost 1.7, LightGBM 4.3, and scikit-learn Random Forest. Foundation models include TabPFN-1.0 (28M parameters) and TabICL-base (100M, LLaMA-style).

TabPFN is restricted to \(\leq\) 10000 rows. We use a fixed random 10k sample for training on Adult, Higgs, and Housing. TabICL and tree models see the full data.

\subsection{Tuning}
Tree models are tuned using 15-trial randomized search with stratified 3-fold cross-validation. Foundation models are evaluated zero-shot with no gradient updates.

\subsection{Hardware \& Instrumentation}

All models run on a single NVIDIA T4 (15GB VRAM) with 2 vCPUs and 13GB RAM (hosted on Kaggle). We record four inference-time metrics: Accuracy (proportion correct on full test set), Latency (wall-clock time per test batch), RAM (peak usage via \texttt{psutil}), and VRAM (peak allocated via \texttt{torch.cuda.max\_memory\_allocated()}).

\subsection{Statistical Testing}
We use the Friedman test on accuracy ranks across datasets, followed by Nemenyi post-hoc test ($\alpha = 0.05$). Hardware cost ratios are computed relative to XGBoost.

\section{Results \& Discussion}

\subsection{Predictive Accuracy}

Table~\ref{tab:acc} summarises test accuracy across the four datasets. Tree-based models consistently perform well: XGBoost and LightGBM both achieve 87.45\% accuracy on Adult, and over 91\% on Housing. Random Forest trails slightly but remains competitive. 

\begin{table}[H]
\centering
\caption{Test-set accuracy (\%).}
\label{tab:acc}
\begin{tabular}{lcccc}
\toprule
Model & Adult & Higgs & Housing & Wine \\
\midrule
XGB (tuned)     & 87.45 & 72.64 & 91.18 & 89.18 \\
LGBM (tuned)    & 87.45 & 72.47 & 91.35 & 88.47 \\
RF (tuned)      & 86.50 & 72.02 & 89.92 & 89.49 \\
TabPFN$^\dagger$ & 85.97 & 71.36 & 91.84 & 88.88 \\
TabICL          & 85.74 & \textbf{73.29} & 91.64 & \textbf{90.00} \\
\bottomrule
\end{tabular}
\end{table}

\vspace{-0.5em}
{\footnotesize $^\dagger$TabPFN trained on a 10\,k-row subset due to context limit.}

Among the foundation models, TabICL delivers the highest scores on Higgs (73.29\%) and Wine (90.00\%), while falling behind XGBoost by 1.7 percentage points on Adult. TabPFN, evaluated on a 10\,000-row training subset due to architectural constraints, stays within one percentage point of the top tree baseline on Housing and Wine, but underperforms slightly on Adult and Higgs.

Across all tasks, tuned gradient-boosted trees remain highly competitive, and the absolute accuracy gains from zero-shot models are modest—typically under 1 percentage point.

\subsection{Inference Cost}

Predictive gains must be contextualised by hardware cost. 

\begin{figure}[H]
\centering
\includegraphics[width=\columnwidth]{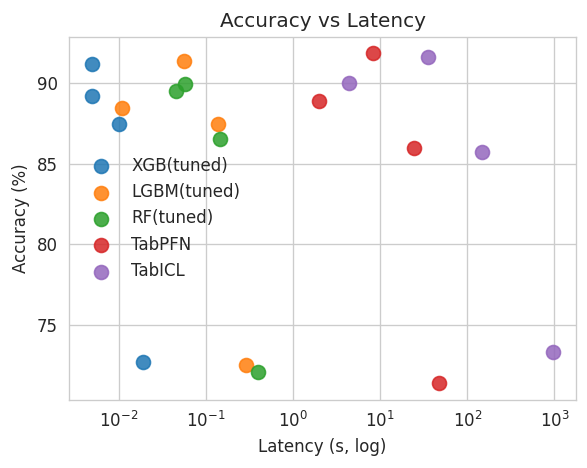}
\caption{Accuracy vs.\ latency (log scale). Classical models occupy the Pareto frontier.}
\label{fig:latency}
\end{figure}

\begin{figure}[H]
\centering
\includegraphics[width=\columnwidth]{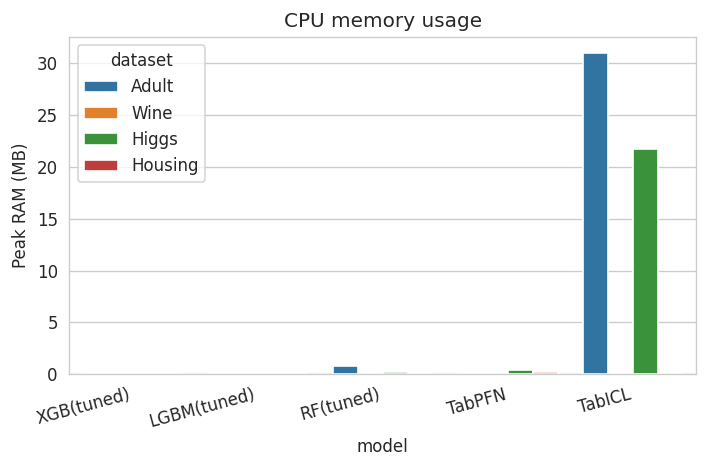}
\caption{Peak CPU RAM usage per model and dataset. TabICL incurs higher overhead.}
\label{fig:ram}
\end{figure}

Figure~\ref{fig:latency} plots accuracy versus latency (log-scale), revealing a clear Pareto frontier. Tree models complete inference in under 0.4 seconds, with XGBoost answering full test batches in as little as 5 to 19 milliseconds. LightGBM and Random Forest follow closely, despite not being GPU-accelerated.

\begin{figure}[H]
\centering
\includegraphics[width=\columnwidth]{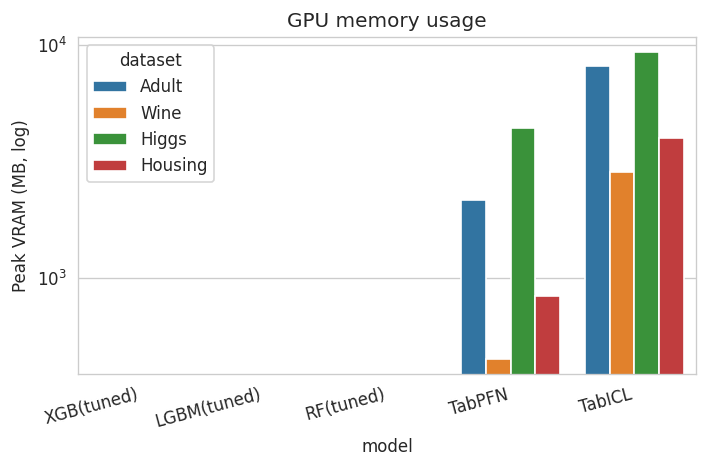}
\caption{Peak GPU VRAM usage per model and dataset. TabICL demands significantly more memory.}
\label{fig:vram}
\end{figure}

In contrast, TabPFN takes up to 47 seconds per batch on Higgs—over 2\,000× slower than XGBoost—while TabICL incurs 960 seconds (16 minutes), representing a 10\,000× latency penalty. These models also consume significantly more GPU memory, detailed in Figures~\ref{fig:ram} and~\ref{fig:vram}.

Figure~\ref{fig:ram} shows CPU RAM usage is negligible across all models except TabICL, which consumes over 20\,MB. While minor in absolute terms, this still exceeds other models by an order of magnitude.

Figure~\ref{fig:vram} highlights peak GPU memory usage. Tree models use no VRAM; TabPFN peaks between 0.8 and 4.4\,GB depending on dataset, and TabICL consistently exceeds 8\,GB. On the Higgs dataset, TabICL reaches 9.3\,GB, exhausting the GPU capacity of most commodity accelerators.

Table~\ref{tab:cost} quantifies median latency and memory multipliers, relative to XGBoost. TabPFN’s inference is over 2\,000× slower, and TabICL is 11\,000× slower, while using nearly 9\,GB of GPU memory. In contrast, LightGBM and Random Forest remain within an order of magnitude of XGBoost latency, with no VRAM usage.

\begin{table}[H]
\centering
\caption{Median inference cost (relative to XGBoost).}
\label{tab:cost}
\resizebox{\columnwidth}{!}{%
\begin{tabular}{lccc}
\toprule
Model & Latency ($\times$) & RAM (MB) & VRAM (MB) \\
\midrule
LGBM (tuned) & 13     & 0   & 0     \\
RF (tuned)   & 18     & 0   & 0     \\
TabPFN       & 2.1e3×   & 0   & 2200  \\
TabICL       & 1.1e4x  & 21  & 8900  \\
\bottomrule
\end{tabular}%
}
\end{table}

These cost differentials are stark: TabICL consumes roughly 35 seconds of additional inference time for every percentage-point of accuracy it gains over XGBoost. This undermines its practical viability for most real-world applications.

\subsection{Statistical Significance}

To determine whether accuracy differences are meaningful, we conduct a Friedman test on model rankings across the four datasets. TabICL ranks highest on average (2.25), with XGBoost close behind (2.62), followed by LightGBM (3.12), and a tie between TabPFN and Random Forest (3.50). The Friedman test statistic is $\chi^2 = 1.97$ with $p = 0.74$, indicating no statistically significant differences. The Nemenyi critical difference is 3.05; all pairwise gaps fall well below this threshold.

This result confirms that while foundation models may occasionally outperform classical baselines on individual tasks, these gains are not statistically robust across diverse settings.

\subsection{Implications}

The results offer three clear takeaways. First, tree-based models such as XGBoost and LightGBM remain the most efficient choice for medium-scale tabular inference, achieving equal or superior accuracy while operating at three to five orders of magnitude lower hardware cost. Second, TabPFN’s fixed 10\,k-row context limit poses a scalability ceiling, forcing subsampling even on moderately sized datasets. Third, although TabICL achieves high accuracy in some tasks, its extreme inference latency and memory requirements prevent deployment in latency-sensitive or memory-constrained environments.

These findings suggest that, in their current form, zero-shot tabular foundation models are best suited for rapid prototyping or low-stakes experimentation on small datasets. Realising their full potential in production workflows will require significant efficiency improvements—via quantisation, distillation, or architectural redesign—or hybrid pipelines where FMs generate high-quality features for efficient classical learners.

\section{Conclusion}

This study provides the first controlled benchmark comparing zero-shot tabular foundation models (FMs) to tuned decision tree ensembles, with accuracy and hardware cost measured in a unified setting. Across four public classification datasets:

\begin{itemize}
\item No statistically significant accuracy difference is observed ($p = 0.74$); TabICL, XGBoost, and LightGBM trade top ranks within ±1 pp.
\item Tree baselines complete full-batch inference in under 0.4s with zero VRAM, maintaining strong performance at minimal cost.
\item TabICL achieves marginal accuracy gains on Higgs, but incurs 960s latency and 9.3GB VRAM, over 10000 times slower than XGBoost.
\item TabPFN matches tree accuracy on small tables but cannot scale past 10k rows due to architectural limits.
\end{itemize}

These results reaffirm that tuned GBDTs remain the Pareto-optimal solution for medium-scale tabular tasks, outperforming current zero-shot FMs in efficiency by orders of magnitude. While FMs may aid rapid exploration on small datasets, their cost profile precludes real-time or resource-constrained deployment.

\textbf{Future directions.} Scaling tabular FMs may require (i) lightweight variants through quantisation or distillation, and (ii) hybrid pipelines that combine FM-driven feature extraction with tree-based inference. Code, data splits, and full benchmark logs are available from the authors upon request to support reproducible, hardware-aware evaluation of structured-data foundation models.

\section*{Impact Statement}
This work quantifies the hardware cost of zero-shot tabular foundation models. It helps researchers and practitioners make informed decisions when choosing between FMs and classical baselines. The findings discourage premature deployment of tabular FMs at scale, and point toward the need for lighter, more efficient architectures.

\bibliography{refs}
\bibliographystyle{icml2025}
\appendix
\onecolumn
\section{Full Inference Cost Tables}

\subsection{Latency (seconds)}

\begin{table}[H]
\centering
\caption{Wall-clock inference latency (seconds) per model and dataset.}
\label{tab:latency-full}
\begin{tabular}{lcccc}
\toprule
Model & Adult & Higgs & Housing & Wine \\
\midrule
XGB (tuned)     & 0.010 & 0.019 & 0.005 & 0.005 \\
LGBM (tuned)    & 0.139 & 0.288 & 0.056 & 0.011 \\
RF (tuned)      & 0.147 & 0.399 & 0.057 & 0.045 \\
TabPFN          & 24.300 & 47.167 & 8.260 & 1.971 \\
TabICL          & 148.237 & 960.634 & 35.666 & 4.411 \\
\bottomrule
\end{tabular}
\end{table}

\subsection{Peak CPU RAM (MB)}

\begin{table}[H]
\centering
\caption{Peak RAM usage (MB) during inference.}
\label{tab:ram-full}
\begin{tabular}{lcccc}
\toprule
Model & Adult & Higgs & Housing & Wine \\
\midrule
XGB (tuned)     & 0.0 & 0.0 & 0.0 & 0.0 \\
LGBM (tuned)    & 0.0 & 0.0 & 0.0 & 0.0 \\
RF (tuned)      & 0.8 & 0.2 & 0.0 & 0.0 \\
TabPFN          & 0.0 & 0.4 & 0.2 & 0.0 \\
TabICL          & 31.0 & 21.7 & 0.0 & 0.1 \\
\bottomrule
\end{tabular}
\end{table}

\subsection{Peak GPU VRAM (MB)}

\begin{table}[H]
\centering
\caption{Peak GPU VRAM usage (MB) during inference.}
\label{tab:vram-full}
\begin{tabular}{lcccc}
\toprule
Model & Adult & Higgs & Housing & Wine \\
\midrule
XGB (tuned)     & 0.0 & 0.0 & 0.0 & 0.0 \\
LGBM (tuned)    & 0.0 & 0.0 & 0.0 & 0.0 \\
RF (tuned)      & 0.0 & 0.0 & 0.0 & 0.0 \\
TabPFN          & 2173.2 & 4402.2 & 838.6 & 450.5 \\
TabICL          & 8181.6 & 9323.1 & 4001.7 & 2844.2 \\
\bottomrule
\end{tabular}
\end{table}

\end{document}